\documentclass[10pt,twocolumn,letterpaper]{article}

\usepackage{iccv}
\usepackage{times}
\usepackage{epsfig}
\usepackage{graphicx}
\usepackage{amsmath}
\usepackage{amssymb}

\usepackage{soul}
\usepackage{url}

\usepackage{booktabs}
\usepackage{algorithm}
\usepackage{algorithmic}
\usepackage{enumerate}
\usepackage{array}
\usepackage{threeparttable}
\usepackage{multirow}
\usepackage{subfigure}

\usepackage[breaklinks=true,bookmarks=false]{hyperref}

\iccvfinalcopy 

\def\iccvPaperID{****} 

\ificcvfinal\fi

\begin{document}

\title{Deep Learning based Wearable Assistive System for Visually Impaired People}

\author{Yimin Lin, Kai Wang, Wanxin Yi, Shiguo Lian\\
AI Department, CloudMinds Technologies Inc., Beijing, 100102, China\\
{\tt\small anson.lin,kai.wang,warsin.yi,scott.lian@cloudminds.com}}

\maketitle
\ificcvfinal\thispagestyle{empty}\fi

\begin{abstract}
In this paper, we propose a deep learning based assistive system to improve the environment perception experience of visually impaired (VI). The system is composed of a wearable terminal equipped with an RGBD camera and an earphone, a powerful processor mainly for deep learning inferences and a smart phone for touch-based interaction. A data-driven learning approach is proposed to predict safe and reliable walkable instructions using RGBD data and the established semantic map. This map is also used to help VI understand their 3D surrounding objects and layout through well-designed touchscreen interactions. The quantitative and qualitative experimental results show that our learning based obstacle avoidance approach achieves excellent results in both indoor and outdoor datasets with low-lying obstacles. Meanwhile, user studies have also been carried out in various scenarios and showed the improvement of VI's environment perception experience with our system.
\end{abstract}

\section{Introduction}

\begin{figure}[t]
\centering
\includegraphics[scale=0.3]{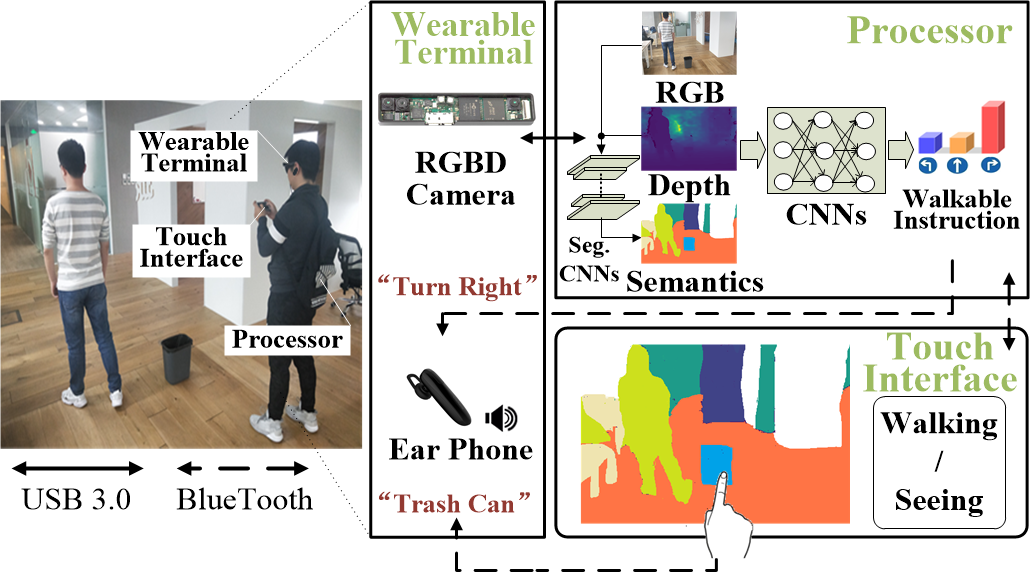}
\caption{VI assistive system overview. Our wearable system includes a wearable terminal consists of an RGBD camera and an earphone, a processor and a touch interface, which provides a walkable instruction and scene understandings to the VI in an efficient way.}
\label{fig:fig1}
\end{figure}

The capabilities of environment perception are crucial for VI~\cite{leo2017computer,tapu2018wearable,bai2019wearable}, such as safe navigation without collision, understanding the room layout or traffic surroundings and object searching in complex and unknown 3D environments. For instance, Fig.~\ref{fig:fig1} shows a specific case in VI's daily life that he is highly expected to safely navigate while avoiding collisions, even with low-lying obstacles (e.g., small trash can) in front of him, and perceives the 3D environment including person, object and backgrounds without complicated interaction.

Specifically, for obstacle avoidance task in robotics community, some sensors (e.g., Lidar, ultrasonic or range sensor) are widely adopted to detect surrounding obstacles scattered in certain distances around the robot, and various conventional path planners can then be employed to instruct the robot along traversable trails~\cite{aman2016sensor-range,flores2018cooperative-lidar}. The major limitation is that the sensors are required to be mounted on a fixed platform without changing in height and direction to guarantee the obstacle detection accuracy. However, it is difficult to employ for human case since the height and direction of the sensor mounted on his head are changing randomly when he is walking, which makes obstacle detection more difficult~\cite{bai2017smart}. Furthermore, to our knowledge, recent approaches have rarely offered a valid way to detect low-lying obstacles (such as curbs and small trash can etc.) that are threats to VI's walking~\cite{yang2018long}.

It is extremely difficult for VI to well understand their surroundings as well as the things they are interested in. OrCAM~\cite{orcam} provides the means to get some special information such as reading texts and detecting traffic lights. However, these information are not enough for VI to understand their surroundings, e.g., they want further to know the room layout and the traffic surroundings. For this, there are two challenges. Firstly, it is not an easy task to detect and describe the plenty of information in complicated surroundings. Meanwhile, few works have been proposed to discuss what is the efficient way for VI to perceive such rich information in 3D environment so far.

Recently, a growing amount of success has been reported for vision navigation and semantic segmentation tasks based on deep end-to-end learning networks~\cite{ortis2017organizing,loquercio2018dronet,chen2018deeplab,leo2018deep,lin2018deep,wang2018unified,han2019deepvio}. Inspired by these works, this paper explores a data-driven learning approach to predict the walkable instruction with supervised learning over training images collected from an RGBD sensor. As shown in Fig.~\ref{fig:fig1}, besides the raw RGB and depth data, we also employ semantic labels as complementary to appearance and structure information to train our networks in order to detect low-lying obstacles. Moreover, semantic labels provide a great help to tell the environment information around VI such as object, road, person and so on. Thus, we also design an efficient way to retrieve the semantic information via simple interactive touching operations on the screen of a touch interface. Both collision-free instructions and object information are transmitted to the user via acoustic signals. We also conduct both quantitative and qualitative experiments on obstacle avoidance, and design user studies to validate our system. The key contributions are summarized as follows:

\begin{enumerate}[(1)]
\item We propose a data-driven end-to-end convolutional neural networks (CNNs) to predict collision-free instructions with moving forward, left and right directions from RGBD data and their corresponding semantic map.
\item We design easy-to-use interactions to provide VI with reliable and efficient feedbacks containing both walking instructions for avoiding obstacles and surrounding information for perceiving 3D environments.
\item We collect real obstacle avoidance datasets about VI under both indoor and outdoor environments covering from day to night, especially including various low-lying obstacles and various illustration conditions for quantitative evaluation.
\item We conduct objective evaluations on obstacle avoidance and other perception tasks by some VI in practical life.
\end{enumerate}

\section{Related Work}

\subsection{Visual Assistive Systems}

Up to now, there exists some vision-based assistive systems, which aim to improve perception experience of VI~\cite{elmannai2017sensor,garcia2019uasisi,bauer2019enhancing}. In particular, OrCAM~\cite{orcam} equipped with monocular camera is used to provide some special object information to VI,  such as reading text and face recognition, etc. It is more suitable for persons with low-vision impaired than for blind persons because it cannot provide the surroundings¡¯ layout and obstacle information. Another wearable system~\cite{wang2017enabling} provides a feedback using range sensor and haptic device when there is an obstacle in front of VI. Similar to this,~\cite{bai2017smart} proposes a smart guiding glasses that tells the walkable instruction using ultrasonic and range sensor. However, both of them do not provide scene understanding ability or well-designed interactive interface. Recently, mobile phone with Talkback software is popular in VI's lives. For instance, NavCog3~\cite{sato2017navcog3} presents a smartphone-based navigation assistant to help VI understand surroundings. However, it only outputs predefined landmarks via speech interaction, which is not an efficient way to browse the layout of environments. Generally, people make decisions based on the thorough understanding of their surroundings. Consequently, such current assistive systems are valid only in limited scenarios and are far away from VI's demand.

\subsection{Vision-based Obstacle Avoidance}

Recently, many traditional techniques~\cite{huang2017visual,reyes2019approach} have been reported for obstacle avoidance in robot navigation tasks. Generally, most approaches rely on range sensors to infer maps of the environment and then compute obstacle-free trajectories. However, these methods are not suitable for VI usage because of the challenges caused by dynamic capturing views mentioned above. By adapting traditional robotic methods for VI, the walkable instruction is detected by analyzing the depth map block-by-block and finding the most distant blocks~\cite{bai2017smart}. However, this kind of method usually requires a large number of manually tuned parameters in order to adapt for different height of person, and is difficult to detect low-lying obstacles. Another prevalent alternative is to use deep learning networks, which enables the development of end-to-end learning approaches to predict the navigation instructions directly from the raw sensory data. Such learning-based algorithms usually rely on supervised training data collected from a human expert controls in a real-world environment. However, the model behaved well just in some special path-like environments with a large amount of RGB training images~\cite{smolyanskiy2017toward}. In contrast, VI would walk at more complicated indoor and outdoor environment, where additional sensor (e.g., range sensor) and high level information is needed to ensure navigation. Moreover, low-lying obstacle avoidance is particularly challenging, which is often discarded by existing algorithms. Thus, this paper aims to learn a VI navigator especially for low-lying obstacles from practical datasets collected with an RGBD sensor.

\subsection{Semantic Segmentation}

CNN-based semantic segmentation has become a popular research topic in the computer vision nowadays. For example, some works~\cite{badrinarayanan2017segnet,chen2018deeplab} use encoder-decoder networks to estimate semantic labels from RGB images. Meanwhile, other segmentation networks with higher performance are proposed by combining the RGB and depth images~\cite{hazirbas2016fusenet,jiang2018rednet}. Intuitively, the semantic labels tell the floor, wall, objects and even small low-lying objects (often regarded as obstacles for VI) on floor. Thus, it is possible to exploit semantic segmentation techniques combined with the appearance and structure information (i.e., RGB and Depth image) to potentially improve the classification accuracy for walkable instruction in this paper. Moreover, semantic information also provides high-level information for the VI to understand the room layout or traffic surroundings, and even helps them to find an object in somewhere. However, few works have been proposed to discuss how to transmit such rich information to VI in an efficient way. Inspired by this consideration, we design an efficient interaction method with a touch interface to help VI locate objects situated in front of them.

\section{System Description}

\subsection{System Overview}

The system architecture and main components are illustrated in Fig.~\ref{fig:fig1}.

\subsubsection{Wearable terminal}


The wearable terminal consists of sunglasses, stereo-based RGBD camera (e.g., Inuitive M3.2~\cite{Inuitive}) and Bluetooth earphone. The RGBD camera can provide highly accurate depth data in a wide variety of indoor and outdoor environments. It provides RGB and depth images both at a resolution of 640$\times$480 pixels at 30 frames per second. The output of the instruction and object information will be translated as audio messages through the earphone to the user. The total weight of the wearable terminal is no more than 150 g.

\subsubsection{Processor (e.g., Laptop)}

In order to satisfy a real-time performance, we chose a laptop containing powerful CPU (Intel I7 8700K) and GPU (NVIDIA GeForce GTX 1080) as processor. Therefore, the CNNs inferences for segmentation and obstacle avoidance tasks are implemented in it. In future, we will implement it on embedded chips. The laptop receives RGB and depth data from RGBD camera via a USB 3.0 cable. Then the collision-free instructions are resent to earphone as soon as possible. On the other hand, it provides the semantic and depth results to the touch interface via Bluetooth transceiver as user needs retrieval.

\subsubsection{Touch interface (e.g., Smart phone)}

The smart phone serves as the main touch-based interaction device in our system. It firstly installs Talkback software to help VI operate the phone. The user can browse the 3D environment information including object class and distance via interactive touching operations on the screen of the phone. This interaction information is sent to earphone through wireless network and play as a voice feedback.

\subsection{Perception Networks}

\begin{figure}[t]
\centering
\includegraphics[scale=0.45]{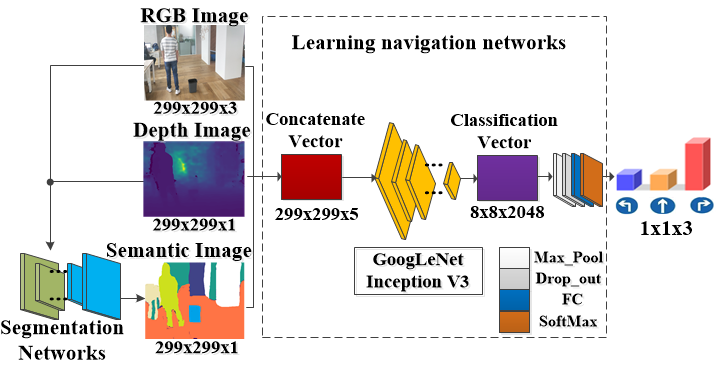}
\caption{Perception networks overview. Segmentation networks provide semantic image using RGB and depth images. While learning navigation networks predict collision-free instructions (turn left, go straight and turn right) using RGB, depth and semantic images.}
\label{fig:fig2}
\end{figure}

As shown in Fig.~\ref{fig:fig2}, the perception networks of our wearable system include segmentation and learning navigation networks.

\subsubsection{Segmentation networks}

We employ FuseNet~\cite{hazirbas2016fusenet} as segmentation networks. It is worth to mention that we firstly retrain the networks using two pixel-level Scannet~\cite{dai2017scannet} and Cityspace~\cite{cordts2016cityscapes} datasets for both indoor and outdoor usages respectively. In particular, 25k training frames with 40 object class labels (i.e. wall, floor, cabinet, bed and chair, etc.) are selected from Scannet, while 5k Cityspace¡¯s fine training frames with 30 classes (i.e. road, sidewalk, person, car and building, etc.) are employed. In addition, we also collect 10k and 4k frames with labels from indoor and outdoor for VI's usage, which are used to fine-tune the networks. Here, we train the indoor and outdoor models with optimal parameters ${C_{{\rm{S1}}}}$ and ${C_{{\rm{S2}}}}$ as follows:

\begin{equation}\label{eq:1}
\mathop {{\rm{arg min}}}\limits_{{C_{{\rm{Sk}}}}} {\rm{ }}\frac{{\rm{1}}}{{\rm{2}}}{\left\| {{C_{{\rm{Sk}}}}} \right\|^{\rm{2}}}{\rm{ -  }}\frac{\lambda }{{mhw}}\sum\limits_{i = 1}^m {\sum\limits_{j = 1}^{h \times w} {\log {p_{{C_{{\rm{Sk}}}}}}({y_{ij}}|{x_{ij}})} }
\end{equation}

where ${x}$ and ${y}$ are the training image and its corresponding ground-truth semantic labels, respectively. ${i}$ and ${j}$ indicate the ${i}$th image and its ${j}$th pixel element. The hyper-parameter $\lambda$ $\textgreater$ 0 is a weight for the regularization of the networks parameters. ${m}$ is the number of training images, and ${h}$$\times$${w}$ represents image size. In order to inference in real time, we resize both of the RGB and depth images to 320$\times$240 and upsample the results with the nearest-neighbor interpolation to the 640$\times$480. The inference runs approximately 25 fps in average in our laptop, which satisfies the real time requirement.

\subsubsection{Learning navigation networks}

\begin{figure}[b]
\centering
\includegraphics[scale=0.16]{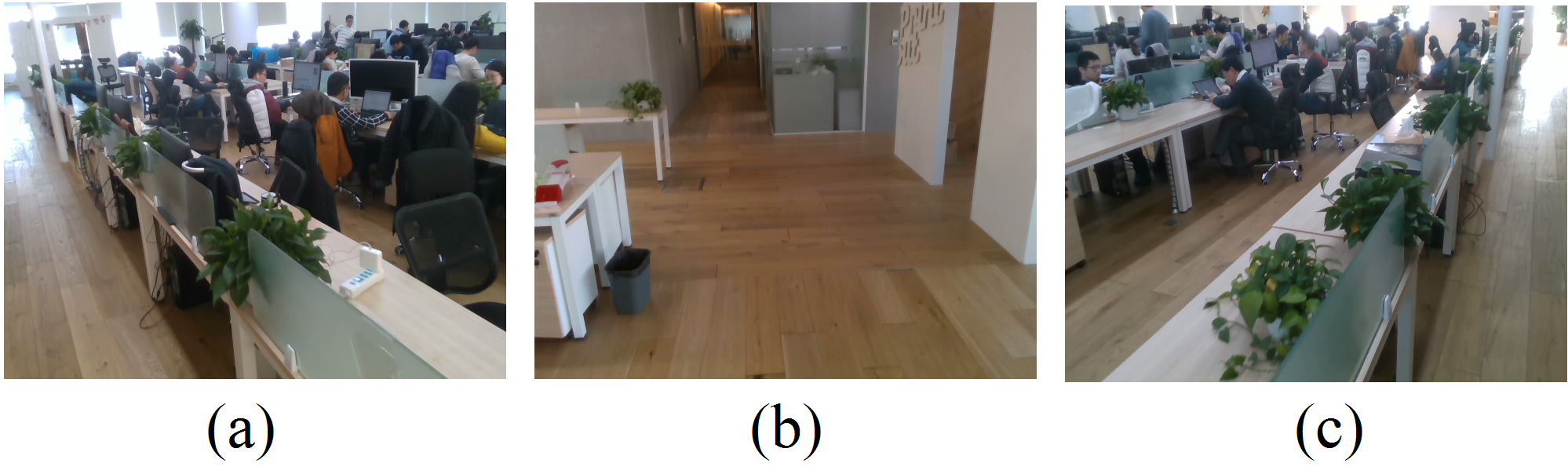}
\caption{Some views from the indoor training datasets related to (a) turn left, (b) go straight and (c) turn right, respectively.}
\label{fig:fig3}
\end{figure}

To our knowledge, there are rare real world RGBD datasets for VI navigation in the literature since collecting such data in reality is difficult and time consuming. Therefore, we design a special way to grasp RGBD data for navigation in real lives. Firstly, a normally-sighted person as pilot, wears our glasses and then walks without collision in both indoor and outdoor environments to collect training datasets composed of RGBD data and three categories of pilot¡¯s actions (i.e. turn left, go straight and turn right). In particular, we capture the datasets when the camera looks to object situated in the right hand side and the action is labeled as ¡°turn left¡±, as illustrated in Fig.~\ref{fig:fig3}(a). Alternately, turn right datasets are captured when looks to another side, e.g. Fig.~\ref{fig:fig3}(c). While the rest of them are labeled as ¡°go straight¡± datasets, as shown in Fig.~\ref{fig:fig3}(b). Totally, we collect 15000 and 6000 annotated images for indoor and outdoor respectively, which also cover day and night scenes under various light conditions. Among them, we highlight that our dataset contains 6000 low-lying obstacle and low-quality depth images, which is rare to consider before. Secondly, we infer the semantic labels from all of them using our segmentation networks.


Finally, these labeled datasets are used to train a model to mimic the pilot¡¯s actions, which is directly known to classify RGBD and semantic images across three categories. Our learning networks are based on standard GoogLeNet but with several changes: we concatenate the RGB, Depth and semantic images as a 299$\times$299$\times$5 input, and add output layers consisting of Maxpool, Dropout and a fully connected layer with three outputs (i.e. probability of turn left, go straight and turn right), activated by a Softmax layer. We utilize binary cross-entropy to train indoor and outdoor models with optimal parameters ${C_{{\rm{N1}}}}$ and ${C_{{\rm{N2}}}}$ through:

\begin{equation}\label{eq:2}
\mathop {{\rm{arg min }}}\limits_{{C_{{\rm{Nk}}}}} \frac{{\rm{1}}}{{\rm{2}}}{\left\| {{C_{{\rm{Nk}}}}} \right\|^{\rm{2}}}{\rm{ -  }}\frac{\lambda }{m}\sum\limits_{i = 1}^m {\log {p_{{C_{{\rm{Nk}}}}}}({y_i}|{x_i})}
\end{equation}

where ${x}$  and ${y}$  are the training input and its corresponding ground-truth action labels, respectively. ${i}$  indicates the ${i}$th image. The hyper-parameter $\lambda$ $\textgreater$ 0 is a weight for the regularization. ${m}$ is the number of training images.
After training, given an input data, the maximum output of the three probabilities indicating the turn left, go straight or turn right is used to instruct VI through collision-free trails. Note that, it runs approximately 10 ms per frame in average in our laptop.

\subsection{Interaction}


\subsubsection{Collision-free navigation feedback}

As we mention above, our system adopts the earphone to provide a text-to-speech feedback. Here, the guiding instruction is sent to VI in real time through a speech interface. Specifically, only two instructions, ¡°turn left¡± and ¡°turn right¡±, are needed to transmit since ¡°go straight¡± is redundant to remind VI. For example, there is no alarm when there are not obstacles in front of VI, thus he can go straight forward. If he receives an alarm from the earphone as ¡°turn left¡± or ¡°turn right¡±, the VI can then move by turning left or right to avoid the obstacle.

\subsubsection{Other perception feedback}

\begin{figure}[b]
\centering
\includegraphics[scale=0.5]{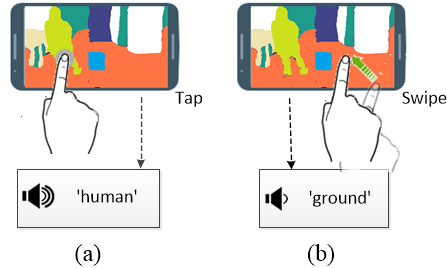}
\caption{The interaction procedure of the proposed system. The user taps or swipes the areas and their class (a) ¡®human¡¯ and (b) ¡®ground¡¯ are played out with different volumes according to their distances.}
\label{fig:fig4}
\end{figure}

This system adopts an easy-to-use interaction design including touch screen and voice play. The first operation is mainly used for sending instructions to actively acquiring the environmental information, and the latter provides feedback of the user¡¯s request.
An example of the detailed interaction procedure is illustrated in Fig.~\ref{fig:fig4}. When the VI stops anywhere and touches the screen to ask for recognition application, then the phone receives a semantic image with depth of the current scene from laptop via wireless transmission. The application will play a ¡¯ding¡¯ sound to indicate the user that the information is ready for retrieval. The user can then touch the screen and swipe around with a single finger to get the information. Each time the finger enters a new region, the name of the object corresponding to this region will be played out through the earphone.
The volume of voice is proportional to the distance of the selected object to the user. The higher the volume is, the nearer the object locates to the user. The current session will stop when the user presses the return key and new sessions are ready to start when the user touches the screen.


\section{Experimental Results }

In this section, we show both quantitative and qualitative results of our obstacle avoidance networks with a set of normal and challenging datasets. Then, we objectively evaluate the performance in guiding and perception for VI in various cluttered environments.

\subsection{Technical Evaluation}

\subsubsection{Training setup}

Our segmentation networks are trained followed the instruction~\cite{hazirbas2016fusenet}. For navigation networks, Gradient Descent Optimizer is employed with up to 100 epochs and the batch size is 16. The learning rate started from 0.001 and decreased by half for every 1/10 of total iterations. We also include Batch Normalization to accelerate convergence and dropout probability is set to 0.2 as a regularization measure. Note that, we utilize 10k indoor datasets for training and the rest of 5k ones are used for testing. While 4k outdoor datasets are used for training and the rest of 2k ones are used for testing. In order to prove that our method generalizes well to new scenarios, test datasets are visually different from the training ones.

\subsubsection{Quantitative results of obstacle avoidance}

To quantify the networks performance on instruction prediction, we use average classification accuracy (ACA) as metric. Four testing scenarios are considered including both Indoor and Outdoor (consists of day and night), Poor depth whose image misses some depth values and Low obstacle which has low-lying obstacles (such as curbs and small trash can etc.). We compare our approach (named as RGBDS) against three baselines: (1) RGB-C is the one of the most state-of-the-art learning approaches who uses RGB as input to predict the three flight-instructions~\cite{chen2018learning}, (2) Depth-T is the traditional work by~\cite{bai2017smart}, and (3) we implement another RGBD-C architecture based on our networks that just removes the semantic labels from the inputs. To our knowledge, there are no learning-based approaches using RGBD data and semantic labels for VI navigation task.

\begin{table}[b] 
	\centering
	\caption{Accuracy results on collision-free instruction classification tasks. (ACA: \%)}
    {
	\label{table1}
	\begin{tabular}{p{1.0cm}<{\centering}p{0.5cm}<{\centering}p{1.1cm}<{\centering}p{1.1cm}<{\centering}p{1.2cm}<{\centering}p{1.2cm}<{\centering}}
		\toprule
		\multicolumn{2}{c}{} &  \footnotesize{RGB-C} & \footnotesize{Depth-T} & \footnotesize{RGBD-C} & \footnotesize{RGBDS} \\
		\midrule
		\multirow{2}{*}{\small{Indoor}}  & \footnotesize{Day} & 92.2 & 94.1 & 95.7 & \textbf{99.6}\\
    \cline{2-6}
             & \footnotesize{Night} & 87.4 & 92.3 & 93.3 & \textbf{98.2} \\
		\hline
        \multirow{2}{*}{\small{Outdoor}}& \footnotesize{Day} & 90.0 & 93.0 & 94.0 & \textbf{98.7} \\
    \cline{2-6}
           & \footnotesize{Night} & 78.1 & 90.0 & 92.3 & \textbf{97.9} \\
        \hline
        \multicolumn{2}{l}{\small{Poor depth}} & 90.2 & 77.1 & 93.8 & \textbf{99.3} \\
        \hline
        \multicolumn{2}{l}{\small{Low obstacle}} & 57.6 & 70.5 & 74.4 & \textbf{98.6} \\

        \bottomrule
	\end{tabular}}
\end{table}

\begin{figure*}[t]
\centering
\includegraphics[scale=0.3]{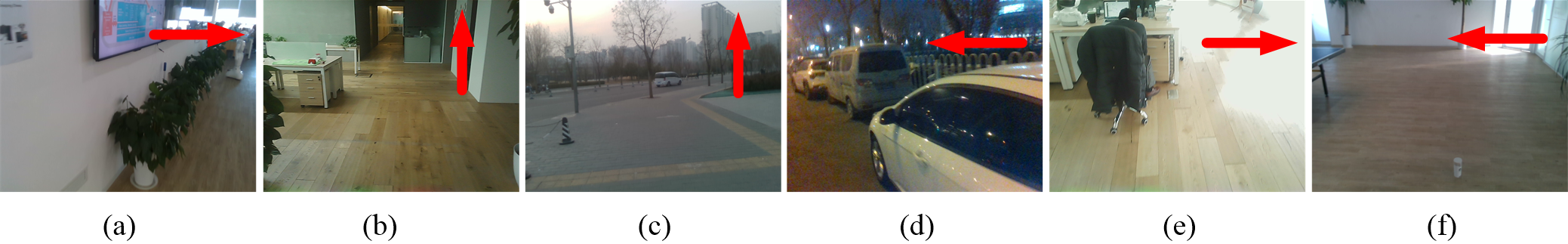}
\caption{Representative samples of our navigation including (a) Indoor day, (b) Indoor night, (c) Outdoor day, (d) Outdoor night, (e) Poor depth and (f) Low obstacle. Red arrow means walkable instruction.}
\label{fig:fig5}
\end{figure*}

Quantitative results are presented in Table 1, the comparison demonstrates that the proposed approach outperforms the three baselines. More specifically, our method has excellent performances in indoor navigation task, where the accuracy achieves 99.6\% and 98.2\% for day and night scenarios, respectively. In contrast, outdoor navigation task is more challenging since it contains big changes in the illumination, visual artifacts or various types of motion. Thus, the accuracy falls to 98.7\% and 97.9\% for day and night scenarios, respectively. For Poor depth datasets, we get a highest accuracy 99.3\% since the segmentation results compensate the missing depth values efficiently. The last Low obstacle tests show that our method pays successful attention to the low object in front of user that results in accuracy 98.6\%. Representative views and results are illustrated in Fig.~\ref{fig:fig5}.

\subsubsection{Qualitative results of obstacle avoidance}
\begin{figure*}[t]
\centering
\includegraphics[scale=0.18]{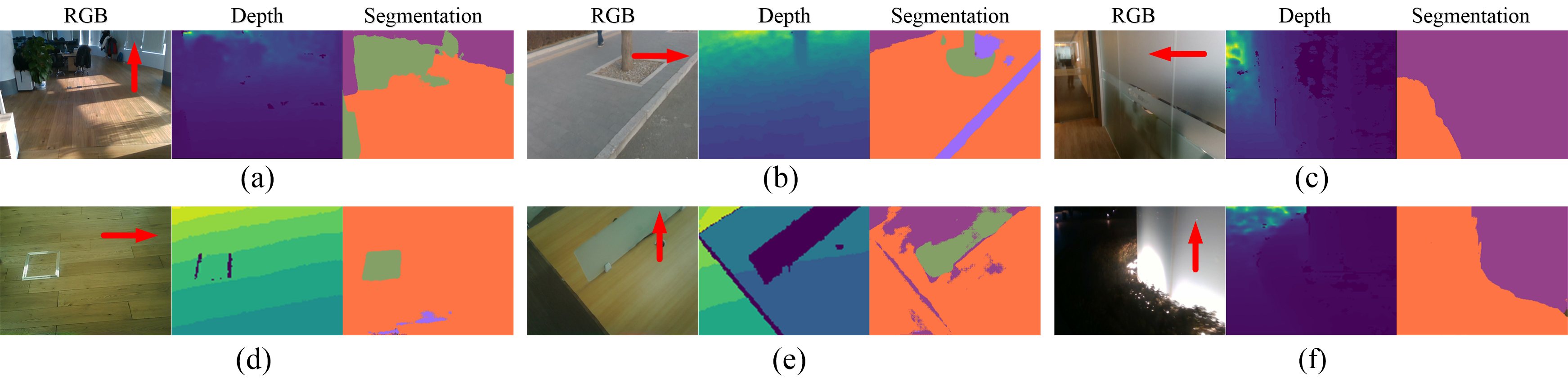}
\caption{Representative samples of our benefits and limitations including (a) various lighting condition, (b) low obstacle, (c)glass wall, (d) poor segmentation, (e) close to obstacle and (f) low lighting condition. Red arrow means walkable instruction.}
\label{fig:fig6}
\end{figure*}

As we all know, vision based perceptions are heavily affected by various lighting conditions, but our method is able to successfully predict a safe instruction even in the case that the sun is shining through the window as illustrated in Fig.~\ref{fig:fig6}(a)-RGB. Moreover, low obstacles are dangerous and common to see in VI's daily lives, such as curbs illustrated in Fig.~\ref{fig:fig6}(b)-RGB. It is difficult to distinguish from raw depth since low obstacles are close to the ground level (as Fig.~\ref{fig:fig6}(b)-Depth). Thus, traditional obstacle detection algorithms often limit a certain height of the object above ground. In contrast, the segmentation (i.e. Fig.~\ref{fig:fig6}(b)-Segmentation) of curb is obvious and beneficial to improve the obstacle avoidance performance in our experiments. A key observation is that depth sensor always comes across failure at glass wall or door, which is common to see in indoor places, as illustrated in Fig.~\ref{fig:fig6}(c)-Depth. However, our method is able to learn features from the segmentation that help it avoid influence from glass walls or doors.

Although our method has been proved considerable potential for VI navigation, some inherent drawbacks need to be overcome. Firstly, the performance is not sufficiently accurate for poor textured regions since the framework is dependent on the quality of the estimated segmentation, as shown in Fig.~\ref{fig:fig6}(d). Secondly, when walking very close to an object or wall, it lacks the perspective view needed to estimate the object label and the framework shows its limitations, see in Fig.~\ref{fig:fig6}(e). Finally, Table 1 indicates that our performance is inferior at night, and Fig.~\ref{fig:fig6}(e) is an example acquired at night. We observe that the illumination condition is poor at the left bottom resulting in a wrong instruction.

\subsection{User Studies}

\subsubsection{Navigation tasks}
\begin{figure}[b]
\centering
\includegraphics[scale=0.25]{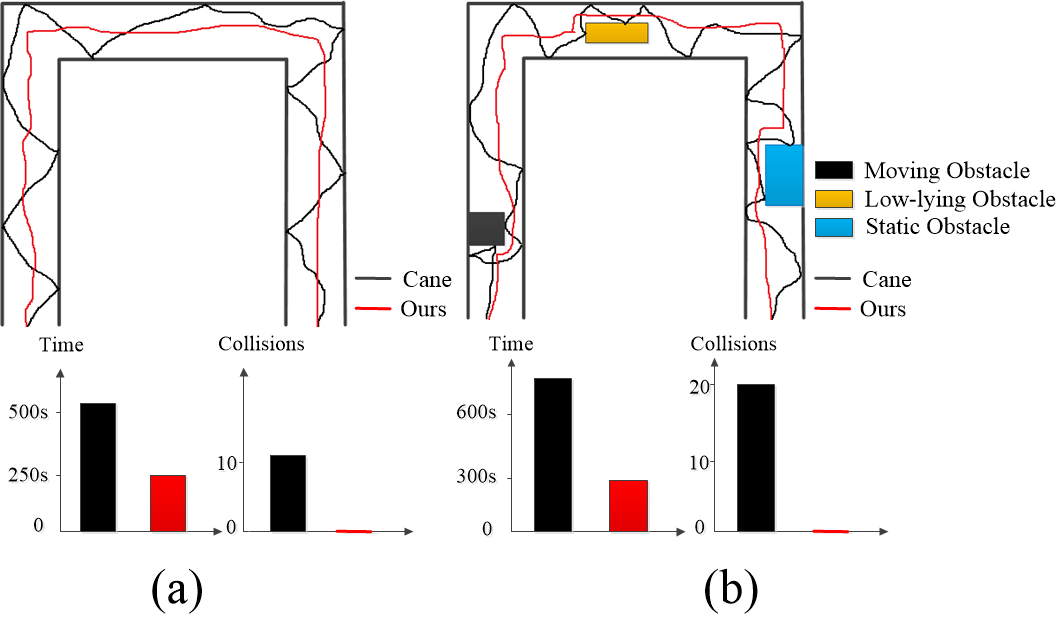}
\caption{VI Navigation studies on (a) hallway with turns and (b) hallway with several obstacles. Walking trajectories are recorded with cane (black) and our system (red), respectively. Comparative metrics are the duration until completion and the number of wall collisions.}
\label{fig:fig7}
\end{figure}

Our system can help users complete safe navigation tasks that they normally perform with the white canes. To accomplish this, we validated our solution through user studies with 20 VI who were totally blind. In the first test, the VI navigated through a 100m long hallway using either a white cane or our system, as shown in Fig.~\ref{fig:fig7}(a). The second test was similar to the first one, but included moving, low and static obstacles situated within the hallway, as shown in Fig.~\ref{fig:fig7}(b). Their walking trajectories were recorded using a Google Tango tablet running a built-in Visual Odometry APP~\cite{Tango2016google}. Here, trajectories for one representative VI's performance are illustrated. The comparative metrics are the mean duration until completion and the number of wall collisions with the white cane or VI's body respectively.

In Fig.~\ref{fig:fig7}(a), we can observe from mean performances that VI took an average of 589s to traverse the hallway using the white cane while 245s with our system.  Furthermore, VI made 11 collisions with white cane while 0 collision with our system. While in the second scenario as shown in Fig.~\ref{fig:fig7}(b), both performances were worse than before with white cane, up to 760s and 20 collisions. In contrast, our system achieved 298s duration with 0 collision as well. The navigation studies suggest that our system improves the mobility performance of the VI by at least 58\% than the users using their white canes in the navigation tasks. Moreover, the system is able to detect all obstacles, both stationary and moving, even low obstacles while continuously moving through the hallway. In addition, trajectories generated by our proposed navigation system are close to the ideal trajectories as normal.

\subsubsection{Other perception tasks}

\begin{table}[t]
	\centering
	\caption{Results of user studies on four tasks.}
    {
	\label{table2}
	\begin{tabular}{p{3cm}<{\centering}p{1.5cm}<{\centering}p{1.5cm}<{\centering}p{1.0cm}<{\centering}}
		\toprule
		Tasks &  Traditional &Ours & EE \\
		\midrule
        Navigation & Cane & AI & 70$\%$ \\
		\hline
        Traffic surroundings & Hearings & AI+Hearings & 52$\%$ \\
        \hline
        Object searching & Touchings & AI+Touchings & 57$\%$ \\
        \hline
        Layout parsing & Hearings & AI+Hearings & 65$\%$ \\
        \bottomrule
	\end{tabular}}
\end{table}

Besides the test of the navigation function, we also asked the participants to perform three other tasks to validate the enhancement of the VI's experience on environment perception with our system. These tasks include traffic surroundings, object searching, and environment layout parsing. For the first task, the users were asked to walk for 20 minutes on 10 different streets, and estimate the traffic surroundings (nearby cars, bicycles, pedestrians, etc.), which they used to perform with their own hearings. For the object searching task, each user was led into 5 rooms, including 2 rooms they are familiar with and 3 they have never been before. For the new rooms, they were given 10 minutes to get familiar about the objects through touching in advance. Then they are allowed to take 10 minutes to get a coarse location of the objects with our system and confirm through touching them. Similarly, in the layout parsing test, each participant is allowed to walk in 5 different indoor environments (2 old and 3 new), accompanied by an assistant with normal vision. They were also given 10 minutes to learn the layout of the surroundings in the new places beforehand. Then they were asked to describe the layout to the assistant. Examples of the descriptions include: ¡°the sofa is in front of me¡±, ¡°there is a chair on right of the desk¡±, etc. The assistants would give instant feedback if the description was inaccurate.
After each test, the participant was given 3 options: ¡®better¡¯, ¡®moderate¡¯ and ¡®worse¡¯, to indicate whether our system had improved their experiences over traditional methods. We computed the percentage of the ¡®better¡¯ options over all the results for each test to quantify the experience enhancement and listed it (e.g., EE) in Table 2.

%

\begin{figure}[t]
\centering
\includegraphics[scale=0.35]{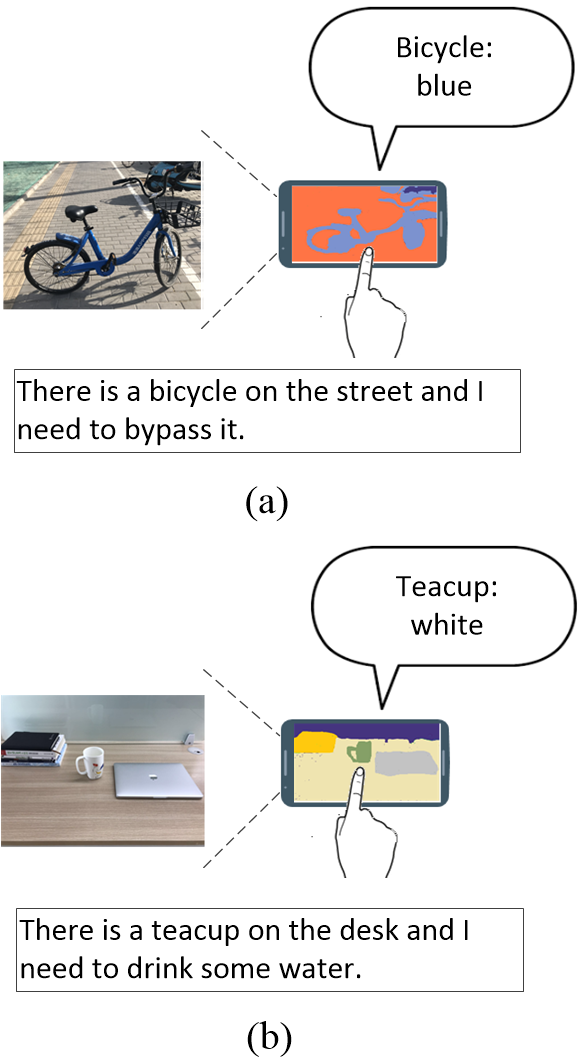}
\caption{Results of VI participants using our system and their feedback. The users were performing (a) traffic surroundings and (b) object searching tasks respectively.}
\label{fig:fig8}
\end{figure}

It can be seen that the enhancement for the navigation task is quite obvious, as the VI participants do not need to use the white cane to explore the obstacles back and forth. By using our system, it also becomes easier to get aware of the object positions and types, as well as the spatial layout of indoor environment. Note that these scores are a little lower than that in the navigation task as the familiar environments which the participants already had a detailed knowledge of were also included, while our system seemed less useful to them in such conditions and 'moderate' or even 'worse' was thus usually selected. The improvement of estimating the surrounding traffic with our system is less obvious. This is mainly because the traffic conditions usually vary within a short period of time, while getting aware of the current conditions through touch screen may result in some delay.
Generally, the users find our system quite useful in performing these tasks, as it indeed helps them complete the operations which they usually need to do in daily lives.  Some examples and the users¡¯ feedback are shown in Fig.~\ref{fig:fig8}.
Through this evaluation, it can be seen that our system works effectively for VI's navigation and object recognition tasks. As one of our participants concludes:
¡®It enables me to walk more freely and know more about the surroundings. I feel like having a pair of powerful eyes when using it!¡¯

\section{Conclusion}


We present a deep learning based wearable system to improve the VI's quality of life. The system is designed for safe navigation and comprehensive scene perception in real time. Specifically, our obstacle avoidance engine, which learns from RGBD, semantic map and pilot¡¯s choice-of-action input, is able to provide safe feedback about the obstacles and free space surrounding the VI. By making use of the semantic map, we also introduce an efficient interaction scheme implemented to help the VI perceive the 3D environments through a smart phone. Extensive experiments on obstacle avoidance demonstrate that our system achieves higher performance in various indoor and outdoor scenarios compared to existing approaches. Furthermore, the user studies on navigation and perception tasks prove that our system improves the mobility performance and environment perception capability in various real scenarios, such as helping to understand the room layout and traffic surroundings, or even helping to find an object at unfamiliar places. Next, the implementation of the algorithms in embedded system will be provided, and the sonar or bump sensor will be incorporated to confirm safety in some extreme case.

{\small
\bibliographystyle{ieee}
\bibliography{ICCV19}
}

\end{document}